\documentclass[letterpaper, 10pt, conference]{ieeeconf}
\IEEEoverridecommandlockouts 
\let\labelindent\relax
\usepackage{enumitem}
\usepackage{cite}
\usepackage{amsmath,amssymb,amsfonts}
\usepackage{algorithmic}
\usepackage{graphicx}
\usepackage{textcomp}
\usepackage{xcolor}
\def\BibTeX{{\rm B\kern-.05em{\sc i\kern-.025em b}\kern-.08em
    T\kern-.1667em\lower.7ex\hbox{E}\kern-.125emX}}

\usepackage[nolist]{acronym}
\usepackage{hyperref}
\usepackage[nameinlink, capitalise]{cleveref}
\usepackage{layouts}
\usepackage{tikz}
\usepackage{pgfplots}
\usepackage{svg}

\crefformat{footnote}{#2\footnotemark[#1]#3}
\usepackage{eso-pic}

\newcommand{\circledchar}[1]{\tikz[baseline=(char.base)]{
    \node[draw, shape=circle, inner sep=0.5pt] (char) {#1};}}

\usepackage{siunitx}      % for typesetting SI units
\usepackage{multirow}     % for multirow cells in tables
\usepackage{caption}      % for customizing captions
\usepackage{booktabs}     % for enhanced table formatting
\usepackage{caption}      % for customizing captions
\usepackage{siunitx}      % Große Zahlen besser darstellen

\usepackage{color, soul}        % TODO: delete after finish, used for highlight text during writing

\AddToShipoutPictureFG*{%
  \AtTextLowerLeft{%
    \put(0,-28){%
      \parbox{\textwidth}{%
        \centering
        \footnotesize
        © 2025 IEEE.  Personal use of this material is permitted.  Permission from IEEE must be obtained for all other uses, in any current or future media, including reprinting/republishing this material for advertising or promotional purposes, creating new collective works, for resale or redistribution to servers or lists, or reuse of any copyrighted component of this work in other works.
      }
    }
  }
}

\begin{document}

\newcommand{\mycomment}[1]{}

\title{\LARGE \bf TUM Teleoperation: Open Source Software for Remote Driving and Assistance of Automated Vehicles}

\author{Tobias Kerbl$^{1}$, David Brecht$^{1}$, Nils Gehrke$^{1}$, Nijinshan Karunainayagam$^{1}$, Niklas Krauss$^{1}$, Florian Pfab$^{1}$, \\
Richard Taupitz$^{1}$, Ines Trautmannsheimer$^{1}$, Xiyan Su$^{1}$, Maria-Magdalena Wolf$^{1}$ and Frank Diermeyer$^{1}$
    % <-this % stops a space
        \thanks{$^{1}$Tobias Kerbl, David Brecht, Nils Gehrke, Nijinshan Karunainayagam, Niklas Krauss, Florian Pfab, Richard Taupitz, Ines Trautmannsheimer, Xiyan Su, Maria-Magdalena Wolf and Frank Diermeyer are with the Institute of Automotive Technology at the Technical University of Munich (TUM), DE-85748 Garching, Germany. Corresponding author: Tobias Kerbl (e-mail: {\tt\small tobias.kerbl@tum.de})}%
        \mycomment{\author{
            \IEEEauthorblockN{1\textsuperscript{nd} Tobias Kerbl}
            \IEEEauthorblockA{\textit{Institute of Automotive Technology } \\
            \textit{Technical University of Munich}\\
            Garching, Germany \\
            tobias.kerbl@tum.de}
            \and
            \IEEEauthorblockN{2\textsuperscript{st}  Frank Diermeyer}
            \IEEEauthorblockA{\textit{Institute of Automotive Technology } \\
            \textit{Technical University of Munich}\\
            Garching, Germany \\
            diermeyer@tum.de}}
        }   
    }
    
    \maketitle

    %% Pre
    \begin{abstract}
Teleoperation is a key enabler for future mobility, supporting Automated Vehicles in rare and complex scenarios beyond the capabilities of their automation. 
Despite ongoing research, no open source software currently combines Remote Driving, e.g., via steering wheel and pedals, Remote Assistance through high-level interaction with automated driving software modules, and integration with a real-world vehicle for practical testing. 
To address this gap, we present a modular, open source teleoperation software stack that can interact with an automated driving software, e.g., Autoware, enabling Remote Assistance and Remote Driving.  %be integrated with the open source automated driving software Autoware, enabling Remote Assistance and Remote Driving. 
The software features standardized interfaces for seamless integration with various real-world and simulation platforms, while allowing for flexible design of the human–machine interface.
% The software features standardized interfaces for seamless integration with real-world systems (e.g., LiDAR, cameras) and operator-side input/output devices (e.g., controllers, head-mounted displays). 
The system is designed for modularity and ease of extension, serving as a foundation for collaborative development on individual software components as well as realistic testing and user studies. 
To demonstrate the applicability of our software, we evaluated the latency and performance of different vehicle platforms in simulation and real-world. 
The source code is available on GitHub\footnote[2]{\url{https://github.com/TUMFTM/teleoperated\_driving/tree/ros2}}.
\end{abstract}

% Teleoperation has emerged as an essential component of future mobility systems, providing support for \acp{av} in rare and complex scenarios that cannot be resolved automated. 
% Despite significant academic research, no up-to-date open source teleoperation software enables Remote Driving, e.g., via control of the steering wheel and pedals and Remote Assistance through high-level interaction with \ac{av}'s software modules. 
% To foster research, this work builds upon a previously released teleoperation software, adding interfaces and functionalities required by current research activities, particularly to enable interaction for Remote Assistance. 
% Our software provides clear interfaces, enhanced visualization capabilities, and further functionality, such as monitoring and logging.  
% To demonstrate the usability and functionality, we apply two teleoperation concepts to our research vehicle. 
% Additionally, we provide a latency analysis of the core data transmission, establishing a baseline for future research. 

% Dadurch können Algorithmen sowie die Mensch-Maschine-Schnittstelle in Versuchsdurchführungen unter möglichst realistischen Bedingungen inklusive der Interaktion mit einem Realfahrzeug untersucht werden.

    %% Main Content
    \section{Introduction}
\label{sec:introduction}
\begin{figure}[!t]
    \centering
    \noindent
    \includegraphics[width=\linewidth]{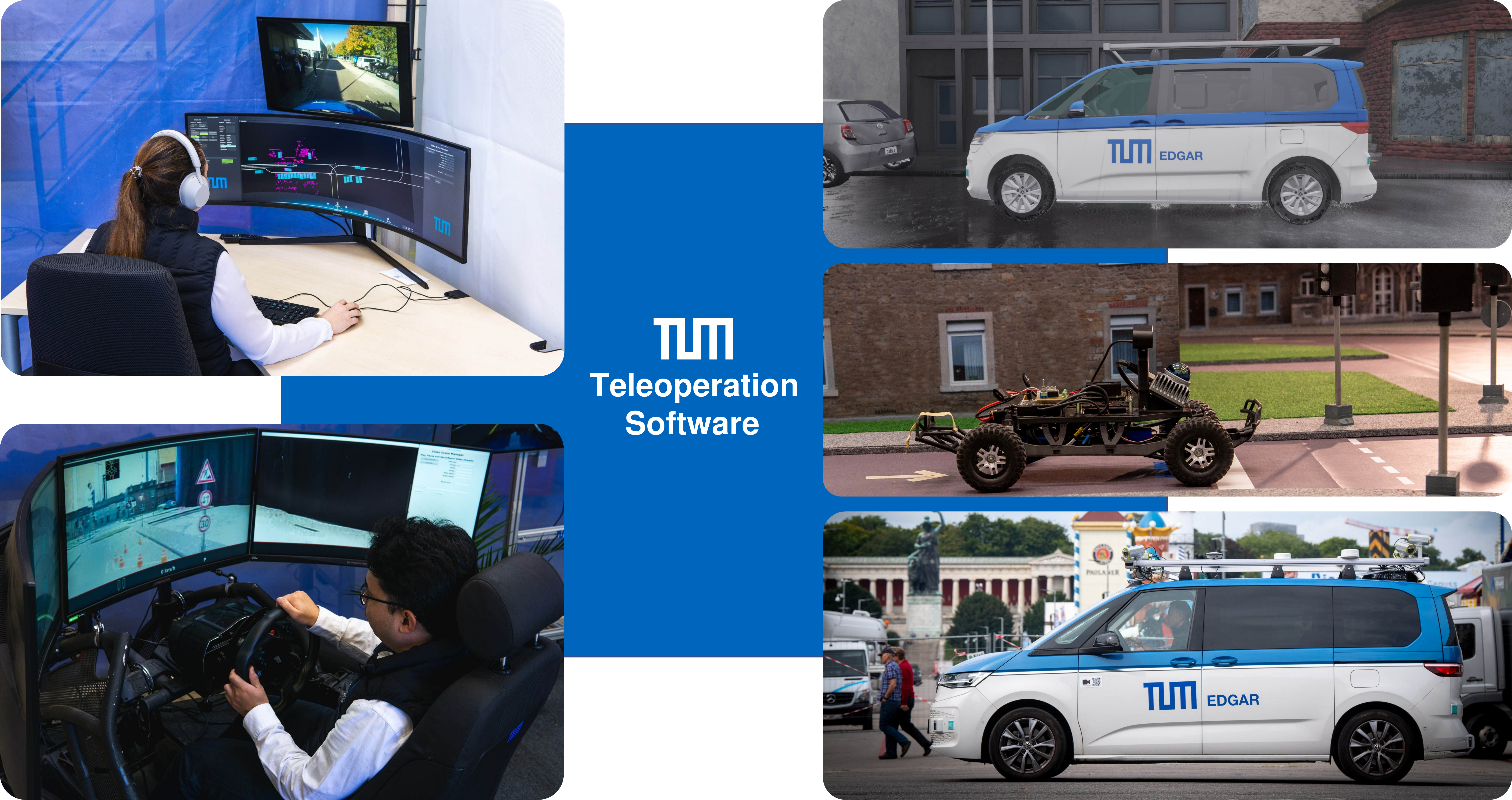} 
    %\includesvg[width=\linewidth]{figures/images/visual_abstract.svg}
    \caption{Teleoperation concepts \textit{Trajectory Guidance} and \textit{Direct Control} demonstrated on teleoperation workstations and research platforms vEDGAR \cite{karle2024}, RoboRacer \cite{roboracer2025}, and EDGAR\cite{karle2024}}
    \label{fig:visual_abstract}
\end{figure}

Teleoperation enables remote support of robots over mobile networks, allowing humans to handle tasks that cannot be fully automated. 
In the field of intelligent vehicles, teleoperation has gained traction, with companies like Fernride and Vay deploying remote driving solutions for logistics and car sharing, gathering significant funding \cite{Fernride2023, Vay2024}. 
Teleoperation also supports \acp{av} during disengagements, as seen with Waymo and Zoox, which rely on \acp{ro} when \acp{av} cannot resolve a scenario~\cite{Waymo2024, NytZooxArticle2024}. 
According to the discontinued company Cruise, in 2024 up to 1.5 people were responsible for operating a \ac{av}, including \acp{ro}  \cite{NytZooxArticle2024}. 
Additionally, regulations such as Germany’s AFGBV \cite{BundesministeriumJustiz2022}, the EU Regulation 2022/1426 \cite{EURegulation2022}, and UNECE standards like Regulation No. 157 \cite{UNECE157} mandate technical supervision and type approval for \ac{av} fleets. 
This demonstrates the demand for teleoperation while highlighting the need for further technological advancements to make it a viable economic solution.
% Although there are many publications and developed systems, there are very few open source software solutions available for vehicle teleoperation, limiting research opportunities.
% For this reason, we publish our teleoperation software developed over the past few years, enabling the control of various vehicle models with two different interaction concepts.
Numerous studies highlight the growing interest in teleoperation. Yet, the absence of open source baseline software forces researchers to repeatedly develop core functionalities, hindering knowledge sharing, reproducibility, and collaboration. A shared open source platform would lower entry barriers, unify development, enable modular improvements, and streamline the creation of functional systems — ultimately fostering more efficient and transparent teleoperation research. % Furthermore, companies mentioned above do not share their software publicly, limiting applicability for research. 
Therefore, this paper presents an open source software stack with an implementation of two distinct teleoperation concepts for \ac{rd} (\cref{fig:visual_abstract}). %, from the \ac{rd} category (\cref{fig:visual_abstract}). 
In \textit{Direct Control}, the \ac{ro} directly controls the vehicle's actors via input of steering wheel and desired velocity. The \textit{Trajectory Guidance} concept allows the \ac{ro} to define a path along with a velocity profile, forming a trajectory that the \ac{av}'s controller is intended to execute.

\subsection{Related Work}

Academic research mainly focuses on individual components of teleoperation systems rather than evaluating complete systems. 
% When entire systems are assessed, they 
Teleoperation solutions are mostly examined using \ac{hmi} click dummies \cite{schrank2024, tener2024}, video recordings without real-world interactions \cite{Brecht2024EvaluationOfConcepts} or custom software that is rarely released or only partially accessible \cite{Housseini2016, gontscharow, Housseini2014, Hoffmann2022, Majstorovic2024trajectoryGuidance}. 
This lack of software accessibility limits knowledge transfer and prevents broader benefits. % from these advancements.

In contrast, companies like Fernride~\cite{Fernride2023}, Vay~\cite{Vay2024} or Zoox~\cite{NytZooxArticle2024} %consistently 
develop 
%and enhance 
their teleoperation capabilities using in-house software. 
Therefore, \ac{rd} and \ac{ra} concepts can be refined and evaluated in real-world settings, but their findings and tools remain proprietary.

To address the limitations of closed source software and foster collaboration in automated driving research, open source projects like Autoware \cite{git_autoware} gained attention, enabling real-world evaluations of \ac{av} software applications, such as in the \textit{Wiesn Shuttle} project \cite{tumWiesn2024}.
However, Autoware primarily focuses on automated functions and offers only a basic teleoperation interface without network or visualization. % components.

To support \ac{rd} evaluation in the open source simulation environment CARLA \cite{dosovitskiy2017}, the ROS-based extension TeleCARLA was developed, integrating video streaming, control transmission, and an \ac{hmi} %remote operator interface 
\cite{hofbauer2020}.
With the aim to enable teleoperation research in real-world applications, a ROS-based open source software stack was developed by Schimpe et al. \cite{schimpe2022}. 
%Its modular design allows for rapid adaption to different vehicle platforms and integration of new functionalities or \ac{av} capabilities. 
It includes video and LiDAR streaming modules, along with network components for efficient data transmission between the vehicle and the \ac{ro}. 
The operator interface provides a \ac{gui} for managing tasks, such as connecting to the vehicle or adjusting video stream parameters. 
As a core element, it includes an \ac{hmi} for \ac{rd}, offering a 3D visualization that features entities such as a vehicle model or video canvas to display vehicle data for teleoperation. 
%Video streams can be projected onto both rectangular surfaces and spherical canvases. 
For operator input, the software supports multiple devices, including steering wheels, controllers, and a virtual joystick. 
Based on this software, valuable insights into different aspects of teleoperation systems, such as teleoperation concepts, \ac{hmi} design, and safety mechanisms, have been obtained, leading to important findings \cite{SchimpeDiss2024, GeorgDiss2024, FeilerDissertation2023, Hoffmann2022}.

Since the release of Schimpe et al.'s software \cite{schimpe2022}, research has shifted from \ac{rd} to \ac{ra} applications \cite{Aramrattana2024}, emphasizing the need for deeper integration with \ac{av} software stacks. 
While the original software design considered individual \ac{av} functionalities, it lacks a clear automation interface for full \ac{av} interaction and implementation of \ac{ra} concepts. 
Furthermore, its ROS-based architecture restricts compatibility with current open source software for automated driving like Autoware which is based on ROS~2. 
%Additionally, unlike Autoware.AI \cite{git_auowareai}, which runs on ROS, the latest Autoware Core/Universe version is now based on ROS2 . 
%As a result, the current version of the open source teleoperation software cannot fully utilize modules from the most widely used open source \ac{av} stack, nor can it support \acp{av} using Autoware, making it less appealing for both academic and industry research.
%The existing design lacks a clear automation interface for full \ac{av} interaction, and x. 
Thus, it cannot fully leverage the latest modules of Autoware, reducing its value for current academic and industry research.

\subsection{Contribution}
This paper presents the \textit{TUM Teleoperation Software}, an open source software stack for teleoperation of ground vehicles, supporting both \ac{rd} and \ac{ra} concepts based on ROS~2. The modular design allows easy integration of custom modules and rapid adaptation to different vehicle platforms and \ac{av} software stacks.
The software provides essential teleoperation functionalities, reducing initial development effort and supporting advanced concept development and module-specific research.
This is showcased through two \ac{rd} concepts on three different platforms (\cref{fig:visual_abstract}). 
Given the critical role of latency in teleoperation, a latency analysis is included to support future research. 
By enabling complete system or module evaluations across various platforms, this project aims to boost collaboration and accelerate teleoperation research.

The paper is structured as follows: \cref{sec:requirments} outlines software requirements, \cref{sec:framework} details the TUM Teleoperation Software and its modules, and \cref{sec:concepts} covers included teleoperation concepts. \cref{sec:vehicles} presents the platforms on which the software was applied, while \cref{sec:experiments} describes %the design and results of the 
experiments demonstrating the software’s capabilities.
    \section{Requirements On The Software Design }
\label{sec:requirments}

The primary objective of the presented software is to facilitate research in teleoperation of ground vehicles by providing a foundation for the implementation and evaluation of a wide range of teleoperation systems across different platforms. At the system level, this includes, but is not limited to, evaluation of teleoperation concepts, safety approaches, and \ac{hmi} designs. Beyond system-level evaluation, the software also supports targeted research on individual components within the teleoperation system context. 
% is also intended to enable focused research on individual components within the context of an teleoperation system.

Importantly, the development of a production-ready software stack is not within the scope of this work. As a result, aspects that are critical in production environments, such as cyber-security, are not considered.

% The software focuses on functionality, innovation, and long-term vision.

\cref{tab:requirements} outlines the requirements necessary to achieve the goal of advancing research in teleoperation with the presented software. ROS~2 was chosen as the framework for the software because it aligns with these requirements and is widely adopted within the research community.

%In contrast, ROS is no longer actively maintained and did not fully meet our needs.

\begin{table}[!ht]
    \centering
    \caption{Main Requirements for the software design}
    \begin{tabular}{p{2.9cm}|p{4.7cm}}
        \toprule
        \textbf{Requirement} & \textbf{Description} \\
        \midrule
        \textbf{R-1} \newline Exchangeable Concepts &  Seamless integration and transition between teleoperation concepts with minimal code modifications. \\
        \textbf{R-2} \newline Platform Adaptability & The software must support rapid adaptation to different vehicle and automation platforms. \\
        \textbf{R-3} \newline Customizable \ac{hmi} &The \ac{hmi} must be customizable, supporting various screen layouts, GUI elements, 2D/3D data visualization, and input devices. \\
        \textbf{R-4} \newline Modularity & The software must provide well-defined interfaces with minimal package inter-dependencies and clear separation of functionalities. \\
        \textbf{R-5} \newline Ease of Use & The software should be user-friendly, both for development and deployment. \\
        \bottomrule
    \end{tabular}
    \label{tab:requirements}
\end{table}

%\textbf{RQ-1: Integration and Modularity of Concepts} \\
%(Teleoperation) concepts and components must be easy to integrate and replace. This flexibility allows for the implementation of diverse control paradigms without requiring significant software modifications during transitions. Similarly, network or safety configurations should be adaptable with minimal effort.

%\textbf{RQ-2: Vehicle Independence} \\
%The software should support straightforward adaptation to various vehicle platforms. This capability is crucial for research, as it enables experimentation across different hardware configurations with only minor adjustments.

%\textbf{RQ-3: Interchangeability of Visualization } \\
%The system must facilitate the testing of diverse visualization approaches, including both 2D and 3D representations, as well as varied screen setups. This ensures the software remains adaptable to different visualization needs and experimental conditions.

%\textbf{RQ-4: Modular Design and Clear Interfaces} \\
%The software stack must incorporate well-defined interfaces to allow for seamless module interchangeability. Dependencies between packages should be minimized, with each package focusing on a single core functionality. This modular approach reduces the effort required to replace individual packages and mitigates interference from other modules, thereby fostering a clean and efficient research environment.

    \section{TUM Teleoperation Software}
\label{sec:framework}

\begin{figure*}[!t]
    \centering
    \includegraphics[width=\linewidth]{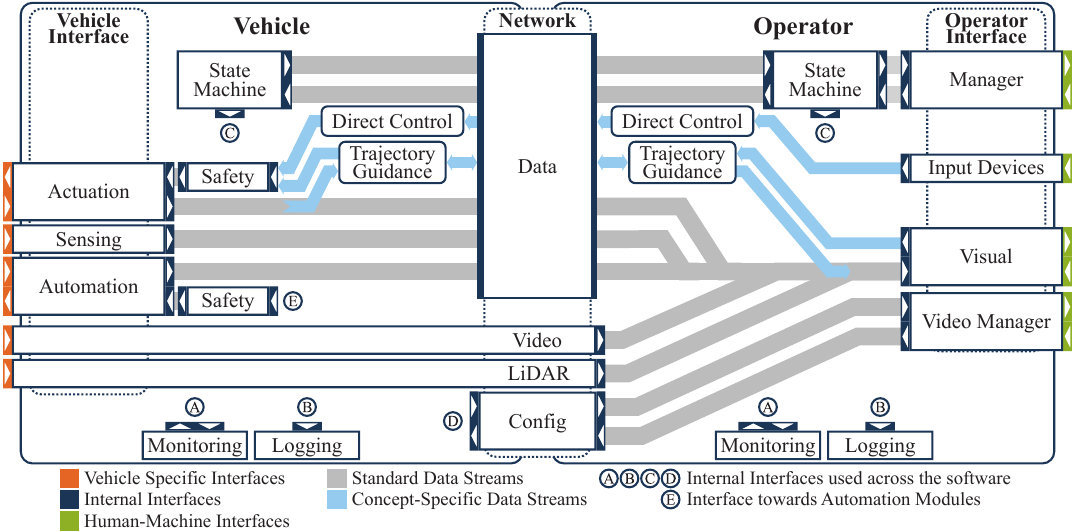}
    \caption{Architecture of the TUM Teleoperation Software with the included teleoperation concepts \textit{Direct Control} and \textit{Trajectory Guidance}}
    \label{fig:architecture}
\end{figure*}

The architecture of the TUM Teleoperation Software is illustrated in \cref{fig:architecture}. The distributed system consists of modules operating both on the \textit{Vehicle} side and the \textit{Operator} side, interconnected through data links. The software is platform-agnostic and can be deployed on any vehicle by implementing a platform-specific \textit{Vehicle Interface}, without requiring changes to the rest of the software (\textbf{R-2}).

To enforce modularity throughout the software (\textbf{R-4}) and ensure that teleoperation concepts can be easily integrated (\textbf{R-1}) the standard functionalities of a teleoperation system (e.g. \textit{Network} or \textit{State Machine} are decomposed into modules with well-defined internal interfaces, including consistent naming conventions and message types. These modules are designed to operate both within the overall software stack and independently, allowing focused research on individual components of the teleoperation system without the need to it entirely (\textbf{R-1}, \textbf{R-3}, \textbf{R-4}). To address the requirement for a customizable \ac{hmi} (\textbf{R-3}) the \textit{Operator Interface} provides base applications (e.g. \textit{Manager}) and supports modular extensions (\textbf{R-4}). 
To simplify the setup and usage, as well as to ensure cross-platform consistency, 
% a consistent development environment across platforms, 
configuration is centralized and a Docker \cite{merkel2014} workflow is provided for development and deployment (\textbf{R-5}). 

Standard data streams, as shown in \cref{fig:architecture},  refer to streams that are largely independent of the specific teleoperation concept being used (e.g., video streaming). The concepts \textit{Direct Control} and \textit{Trajectory Guidance} demonstrate how teleoperation concepts or new functionalities can be integrated using the internal interfaces.

\subsection{Vehicle Interface}
To work with any provided research platform (\textbf{R-2}), a vehicle interface is defined, see left on \cref{fig:architecture}. %.
%Therefore, ROS topic interfaces towards the platform are ensured within the software framework 
% while using external platform-specific interfaces. 
% The purpose of the interface is to ensure ROS topic interfaces towards the platform inside the software framework while using the external platform-specific interfaces.
Experience from past implementations, as presented in \cite{schimpe2022}, revealed that the previous architecture required modifications to both source code and configuration files across multiple locations in order to deploy the software on a new platform. The new architecture addresses this issue by eliminating multi-package dependencies and introducing a generic interface. 

For any new platform, the vehicle interface is divided into two components: a platform-specific interface implementation based on the generic interface and a set of configuration files. The generic interface defines the communication with the teleoperation software, specifying the expected message types and required signals. Researchers are only responsible for implementing conversions from their platform-specific data formats to this generic interface. The interface implementation is organized into the following modules, each with a distinct responsibility:

\begin{itemize}
    \item \textit{Sensing} -- Processing of low-bandwidth data, e.g. vehicle position and accelerations
    \item \textit{Actuation} -- Processing of data from and towards the actuation, e.g. steering wheel position and velocity
    \item \textit{Automation} -- Processing of data from and towards the automation, e.g. automation state, perceived objects or desired trajectory
\end{itemize}

%The vehicle interface is split into a Video  and LiDAR as well as a Sensing interface processing low-bandwidth data like vehicle position or orientation%GNSS and IMU
%, an Actuation interface processing data from and towards the actuation, and an Automation interface processing data from and towards the automation. 
%Based on these interfaces, platform-specific ROS2 packages can be implemented. 

The second component of the vehicle interface, the configuration files, is defined within the centralized launch structure, eliminating the need to update multiple packages individually. These files include platform-specific parameters such as vehicle dimensions and sensor positions.

\subsection{Network}
\label{sec:network}
The network modules (center on \cref{fig:architecture}) serve a significant role in the architecture since they enable bi-directional communication between the vehicle and \ac{ro}. The network component comprises four modules: 

\textit{Video} handles the crucial task of video streaming. To achieve low latency and configurability in video streaming we utilize the open source project GStreamer project with a ROS~2 integration following Schimpe et al.\cite{schimpe2022}. The streams are encoded using the H.264 standard with ultra-low latency settings and transported via separate RTSP channels. The camera streams can be configured using a \ac{gui} in the \textit{Video Manager} to control the bit-rate and the stream status, described in \cite{schimpe2021}. 
It also supports a multiple-router setup with different mobile network providers for redundant streams to increase reliability during sudden package drops.
% Similar work using FFMPEG can be found under \cite{rosFFMPEG2024}.

\textit{LiDAR} uses a modified ROS~2 Point Cloud Transport to compress and transmit the point clouds across the network. The point cloud can be filtered and optionally smoothed. 
% On the EDGAR platform we observe that the filtered concatenated pointcloud results in roughly 50.000 points and transmitted with 10Hz using roughly 300Kbps. 

\textit{Data} provides a centralized package to handle low-bandwidth data via the network. By using ROS~2 generic subscriber and publisher, it achieves low latency transmission for ROS~2 topics. The topic name, type, as well as the ports are easily configurable.

\textit{Config} utilizes ROS~2 services to configure % various 
parameters during run-time. A set of service forwarders and listeners is implemented to bridge ROS~2 services from the \ac{ro} to the vehicle (interface \circledchar{D} in \cref{fig:architecture}).

Both \textit{Data} and \textit{Config} support UDP and TCP and can be easily configured upon start-up (\textbf{R-4, R-5}).
%The service forwarding supports also UDP and TCP protocols currently. 
%However, unlike the sender and receiver, the TCP protocol is considered more suitable for the ROS2 service forwarding, since ROS2 service requires a response to finish the service call. 
%In such cases, the TCP protocol delivers more reliable responses compared to the UDP protocol. 

\subsection{Operator Interface}

The operator interface enables the \ac{ro} to interact with the vehicle and \ac{av} system through an \ac{hmi} and corresponding input devices (right side of \cref{fig:architecture}).

\textit{Visual}: To bridge the gap between open source teleoperation tools and industry solutions like Waymo \cite{Waymo2024}% and Zoox \cite{NytZooxArticle2024}
, an extensible graphics library was developed to support interactive \ac{hmi} design for various teleoperation concepts (\textbf{R-3}). Based on Schimpe et al. \cite{schimpe2022}, the 3D scene renderer integrates an entity-component system, ROS~2, and the \ac{gui} framework Dear ImGui. Rendering layers allow \ac{gui} elements to appear independently of the 3D scene, improving usability and enabling run-time switching between concepts (\textbf{R-1}). %with the vehicle. 
%The combination of these elements enables the application to switch between different teleoperation control concepts during the connection with the vehicle.
% Through the combination of the \ac{gui} framework and the 3D scene rendering, it is possible to integrate various teleoperation concepts. 

The vehicle's state, map location, as well as sensor and perception information, can be displayed within a 3D scene. Depending on the teleoperation concept, the scene can be flexibly configured — for example, to either display an immersive 3D projection of the camera streams in the case of \textit{Direct Control} (\cref{fig:direct_control}) or to abstractly represent the vehicle's environment and realize trajectory-based control in \textit{Trajectory Guidance} (\cref{fig:trajectory_guidance}).

% The \textit{Visual} provides a simple, extensible, and performant way to develop real-time graphics applications for ROS2. 
In comparison to prevalent alternatives such as RViz2 and Foxglove-Studio, which are encumbered by challenges such as integration complexity and limited extensibility, our software provides a distinct advantage in terms of simplicity, extensibility, adaptability, and performance.
%This was investigated in a study with 45 participants in a between subject design in conjunction with different positioning of the elements and input devices. The \ac{gui} now published is strongly influenced by the result of the study findings \textcolor{red}{[CITE WOLF]}.

%\begin{itemize}
%    \item  Changes compared to \cite{schimpe2022} i.e. extension of the ECS and ROS Integration, Control State based Rendering, 
%    \item Extended the the graphics libray via the Graphical User Interface library dearImGui \cite{imGui2024}, Rendering Layers for good UIs (future work), e.g. Displaying Image Streams outside of 3D Scene, Extended GUI support for Interfaces e.g. TG and UILayers
%    \item Allows for pratical UIs in combination with a digital twin for different concpets e.g. TG PM CP 
%    \item Now serves as the central place for the graphical user input e.g. Manager UI and SceneManager UI 
%    \item Usability/User Experience (Feedback from Maria's studys or cite her paper)
%    \item tod gl as easy way to build applications with separation between GUI and ROS

%\end{itemize}
%\todo[inline]{Add graphic manipulation options through the input devices}
%\begin{itemize}
 %   \item input waypoints trajectory
 %   \item set speed
 %   \item steering angle (Fahrschlauch?)
 %   \item turn signals
 %   \item settings in managers
 %   \item viewing/camera angle
 %   \item driving mode
%\end{itemize}

\textit{Manager}: The Operator Manager oversees the teleoperation session setup and coordination. It manages the connection process to vehicles, allows the selection of teleoperation concepts and input devices at run-time (\textbf{R-1}), and handles the initiation and termination of vehicle interaction. Furthermore, it displays key system states such as connection status, selected concept, software health, and other relevant KPIs.
% Created with tod\_gl,our self-created graphics library based on dearImGui, there are also two managers that allow general settings of the operator and to the transmitted videos. Each of these managers opens in its own window and allows the elements to be arranged by docking layer.

%The operator manager is used to connect with the vehicle, change the teleoperation control concept as well as the set the input device used by the operator. Furthermore, it displays network statistics and vehicle interface information during connection to the vehicle.    

\textit{Video Manager}: The video manager, allows the \ac{ro} to change the configuration of the transmitted video streams during run-time \cite{schimpe2021}. Such as activating or deactivating individual video stream, as well as setting the bit-rate or crop the size of each video.

\textit{Input Devices}: 
The software enables the use and easy integration of various input devices, including USB peripherals such as mouse and keyboard or game controllers %, and joysticks
(\textbf{R-3}).

% In addition to controlling the vehicle asynchronously via mouse and keyboard, the software supports to directly drive the vehicle via different USB devices such as a steering wheel and pedals \cite{schimpe}. 

% What matters for the specific teleoperation concept are the data processed by the vehicle, such as for Direct Control the steering angle and acceleration. 
% However, it is essential to note that all control signals required for the respective teleoperation concept can be input through the hardware. 
% The published software allows Direct Control using a Fanatec or Sensorwheel steering wheel and pedals, as well as a virtual input device with a mouse or keyboard keys to steer (W-A-S-D) or change gears (T-G). Trajectory Guidance is also implemented using a mouse and keyboard, building on the interaction concept by Wolf et al. \textcolor{red}{[CITE WOLF]}.

% (For debugging purposes, there is also a virtual input device that can be used for testing without having to connect a hardware input device.)

% In addition to controlling the vehicle asynchronously via mouse and keyboard, the software supports to directly drive the vehicle via different USB devices such as a steering wheel and pedals \cite{schimpe}. 

\subsection{State Machine}

\begin{figure}[!ht]
    \centering
    \begin{center}
        \includegraphics[width=\linewidth]{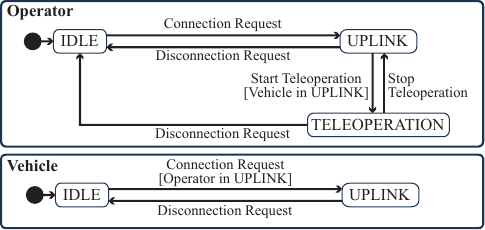}
    \end{center}
    \caption{State diagram of operator and vehicle state machine}
    \label{fig:tod_state_machine}
\end{figure}

To manage and communicate the status of the system between the \ac{ro} and the vehicle, two coupled state machines are employed, with one running on the \ac{ro} side and the other on the vehicle side.
The states and transitions of each state machine are depicted in \cref{fig:tod_state_machine}. 
The operator state machine consists of three states \textit{IDLE} (no connection between vehicle and \ac{ro}), \textit{UPLINK} (only vehicle data is being sent) and \textit{TELEOPERATION} (teleoperation is active). 
The vehicle state machine is a reduced version of the operator state machine with its two states \textit{IDLE} and \textit{UPLINK}. Interface \circledchar{C} facilitates the access to both system's state information.
 %The state transitions in the operator state machine are mainly triggered by the \textit{Manager}, except the transition to the \textit{TELEOPERATION} state, which additionally depends on the vehicle's state. 
%The vehicle state machine changes its state either on request of the operator state machine to \textit{UPLINK} or triggered by the \textit{Manager} to \textit{IDLE}.  

%If a connection is requested via the Operator Manager, the operator state machine transitions to \textit{UPLINK}. This triggers the state transition from \textit{IDLE} to \textit{UPLINK} on the vehicle side, about which the operator state machine is informed. As soon as the teleoperation is started in the Operator Manager, the operator state machine checks the delivered vehicle's status. If it is in \textit{UPLINK}, the operator state machine switches to \textit{TELEOPERATION}. In this state, the teleoperation process can be executed. Once the operator stops the teleoperation process, a state transition from \textit{TELEOPERATION} to \textit{UPLINK} is triggered. If a disconnection is requested, the operator state machine transitions to its initial state. This also causes the state to transition back to the initial state on the vehicle side. 

\subsection{Monitoring}
% Monitoring is essential for ensuring the safety and reliability of automated systems. 
% Similar to Autoware’s monitoring module, the proposed software stack includes a dedicated monitoring component to evaluate current system capabilities. 
% Since the mobile network is a safety critical component in the teleoperation system, a network monitoring module has been implemented to assess link quality, including latency and available bandwidth. 
% Additionally, a topic monitoring module tracks the availability of internal data streams. The monitoring module is designed to be extendable. %, allowing for future extensions.

Continuous monitoring of system capabilities is essential to ensure the safety and reliability of the overall system. To achieve this, a dedicated monitoring component is proposed.
Due to the inherent variability of mobile networks, a network monitoring module is provided. This module is capable of assessing link quality, latency, and available bandwidth.
Additionally, a topic monitoring module observes the availability and consistency of internal data streams. The input and output (interface \circledchar{A} in \cref{fig:architecture}) of the monitoring framework is designed to be modular and extensible (\textbf{R-3}).

\subsection{Safety}
% The safety module's task is to trigger mechanisms that ensure the teleoperation system remains safe based on evaluation of the current system capabilities.
% Mechanisms may include the initiation of a safe stop if the network connection is lost.  
% In degraded states, such as low camera frame rates, the module can limit the vehicle’s maximum speed or restrict teleoperation concepts, such as \textit{Direct Control}, which rely on a stable connection.
The safety module ensures that the actions executed by the vehicle align with its current capabilities. Based on the outputs of other modules (e.g. \textit{Monitoring}, \textit{State Machine}) the safety module determines whether primary and secondary control commands should be forwarded to the vehicle's Actuation, restricted, or overridden to trigger a safe stop. Even though not utilized at the time of publication, commands directed toward the automation are also intended to pass through the safety module (interface \circledchar{E} in \cref{fig:architecture}).
% This mechanism guarantees safe operation in accordance with the system's current state.

\subsection{Logging}
Since debugging options are crucial for development and evaluation of new software components, a logging component is introduced in the software. Utilizing interface \circledchar{B} modules can specify information to be logged (\textbf{R-4}, \textbf{R-5}).
%This component collects all available ROS~2 topics within a given namespace both on the operator and the vehicle side.
%The content of the ROS2 topics is written into a YAML file, providing debug information to the operator or user.  

    \section{Teleoperation Concepts}
\label{sec:concepts}

The interaction between the \ac{ro} and the vehicle can be categorized into different teleoperation concepts. For term definitions, advantages, and disadvantages, see the publications by Majstorovi\'c et al. \cite{Majstorovic2022} and Brecht et al. \cite{Brecht2024EvaluationOfConcepts}. 
The developed software incorporates two Remote Driving concepts: \textit{Direct Control} and \textit{Trajectory Guidance}. 
%Notably, these interactions can be implemented without relying on functional vehicle automation.

\subsection{Direct Control}
\label{sec:direct_control}
% [2 paragraphs, max. 3/4 of one column]

\textit{Direct Control} is the most commonly used teleoperation concept in industry and research.
%Direct Control is a commonly used concept in teleoperation. It is frequently discussed in both research and industry [sources]. 
This approach enables the \ac{ro} to control the vehicle by directly transmitting steering and velocity inputs, just like driving a vehicle from afar.
%This approach allows the operator to control a vehicle in a manner similar to normal driving, where steering and velocity inputs are directly transmitted to the vehicle. 
%%Velocity control is favored as high system latency can complicate acceleration control for human operators \textcolor{red}{[SOURCE]}. 
%Velocity control is preferred because acceleration control can become problematic for human operators when system latency is high [source]. 
% The velocity is later converted into an acceleration input for the vehicle.
\textit{Direct Control} stands out for its controllability and simplicity compared to other teleoperation concepts while being more prone to network instabilities and potentially inducing high mental workload to the \ac{ro} \cite{Brecht2024EvaluationOfConcepts}. 
% However, it carries a high risk of human driving errors and requires high mental workload and network stability. 
%. Therefore, this teleoperation system places significant mental demands on the operator and requires high network stability \textcolor{red}{[SOURCE]}. 
%However, task completion times in Direct Control are shorter than separate input of path and velocity \textcolor{red}{[CITE WOLF]}.

%The advantages of Direct Control include faster task completion times 
% >> faster than what?
%and a more familiar interface compared to trajectory guidance systems. 
% >> why? what if we click on camera view?
%However, its disadvantages include the need for low system latency and a higher workload for the operator.

%As Direct Control is one of the two implemented concepts in this open source software stack, the code 
The software provides interfaces for steering, accelerator and brake pedal positions, along with functionality for creating, sending, and receiving commands. 
The commands are divided into primary (steering and velocity) and secondary control (gear changes, turn signals, honking, etc.) commands. 
%The primary command transmits steering and velocity inputs, while the secondary command handles gear changes, turn signals, honking, and similar functions.

\begin{figure}[!h]
    \centering
    \includegraphics[width=\linewidth]{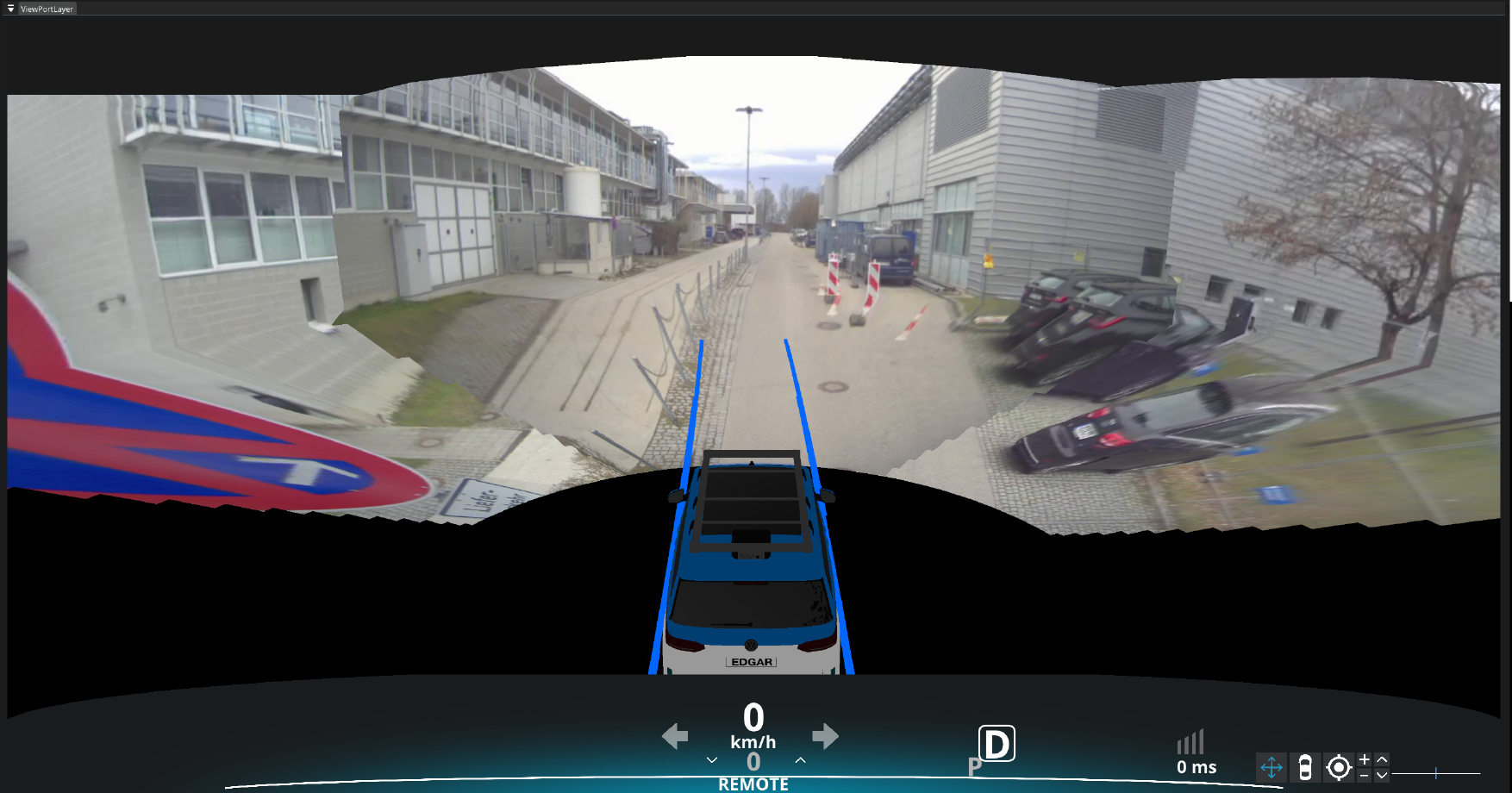}
    \caption{Operator Interface in \textit{Direct Control} mode}
    % \caption{3D Projection of videos in \textit{Direct Control} mode}
    \label{fig:direct_control}
\end{figure}

%To keep the operator informed about the vehicle's state and its environment, a \ac{gui} is implemented. 
In \textit{Direct Control} mode, \textit{Visual} (\cref{fig:direct_control}) displays stitched low-latency video streams projected in the 3D scene and a bottom bar including state information, such as target and actual speed, gear, and network latency, based on Wolf et al. \cite{Wolf2025}. 
Additionally, the system's teleoperation mode is displayed and colored in blue. 
The 3D scene can be adjusted by zooming, rotating, or moving and switching between a third-
person perspective and a bird’s-eye view using the buttons at the bottom right. 
However, the view should remain static during driving to avoid distraction.

% First paragraph (general introduction for DC):
% \begin{itemize}
%   \item DC is the most standard teleoperation control concept
%   \item direct control from research and industry [\hl{considering overlapping with the second chapter}]
%   \item pros and cons for direct control
% \end{itemize}

% Second paragraph (technical details) [\hl{considering overlapping with the third chapter}]
% \begin{itemize}
%     \item input devices (e.g., steering input, velocity input) 
%     \item network (e.g., vehicle data, vehicle state, video stream, and control command are transmitted)
%     \item visual (e.g., fahrschlauch, 3D projection, stitching)
%     \item control (e.g., control commands are aggregated and sent with fixed frequency)
% \end{itemize}

\subsection{Trajectory Guidance}

The \textit{Trajectory Guidance} concept addresses some of the shortcomings of \textit{Direct Control}, including lower mental demand for the \ac{ro} \cite{Wolf2024TrajectoryGuidance} and reduction of adverse effects like latency on the system due to decoupling of the trajectory specification and trajectory following task \cite{Brecht2024EvaluationOfConcepts}. 
Disadvantages comprise higher completion times to solve a given scenario \cite{Majstorovic2024trajectoryGuidance, Wolf2024TrajectoryGuidance} and increased system requirements due to the need for a precise localization. 
%needed to implement this concept.
%TG enables safety concepts based on trajectory level also less effort for the remote operator
%In continous trajectory guidance aka safe corridor (we dont do that here) → Show examples

The developed software enables the \ac{ro} to define trajectories by specifying a series of points in a 3D environment in \textit{Visual} and set the desired velocity for the vehicle, using either mouse and keyboard or a touchscreen \cite{Wolf2024TrajectoryGuidance}. 

\begin{figure}[!ht]
    \centering
    \includegraphics[width=\linewidth]{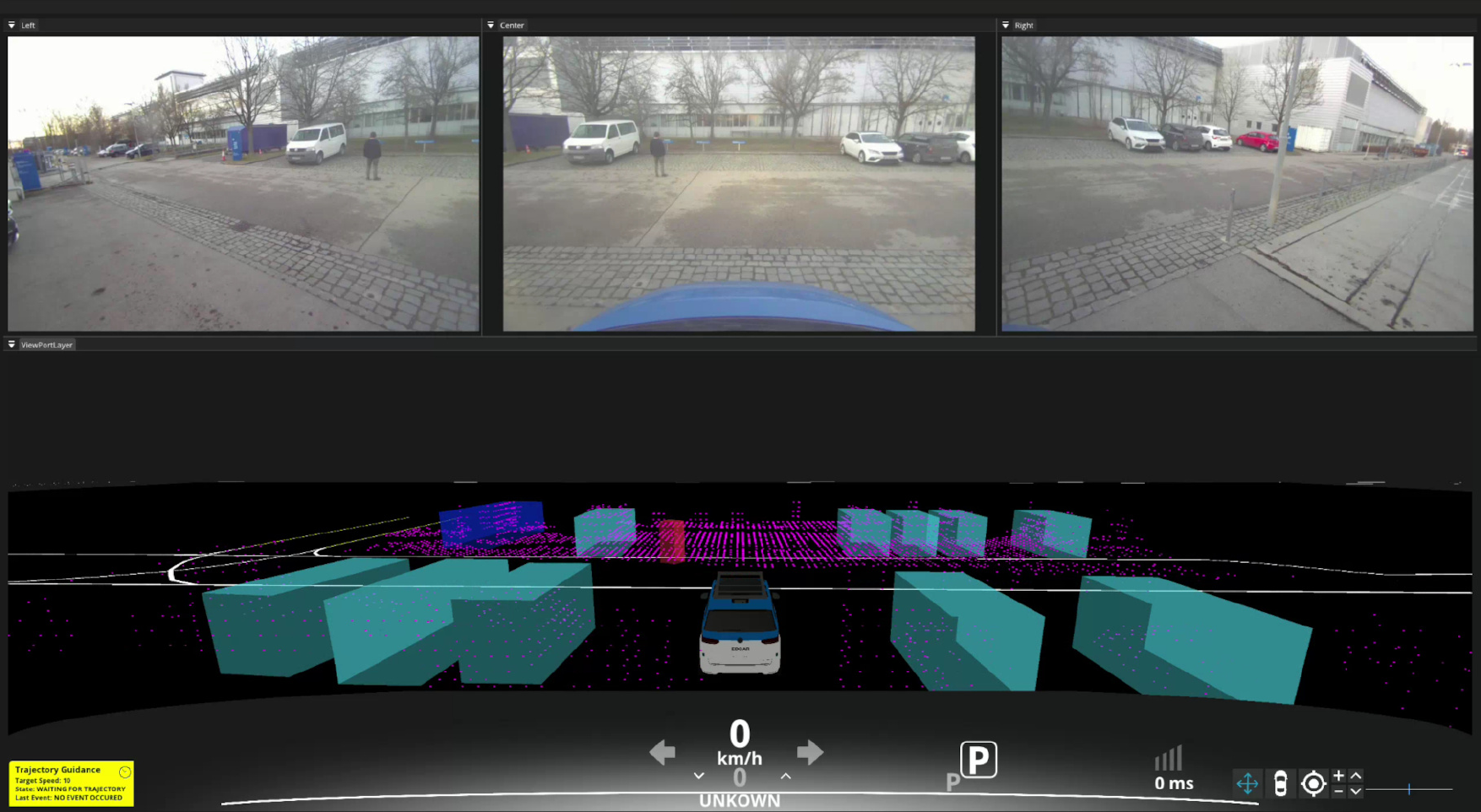}
    \caption{Operator Interface in \textit{Trajectory Guidance} mode}
    \label{fig:trajectory_guidance}
\end{figure}

In \textit{Trajectory Guidance} mode, \textit{Visual} (\cref{fig:trajectory_guidance}) features a horizontal split screen, typical in industrial applications \cite{Cruise2021, Zoox2020}. %and researched by Wolf et al. \cite{Wolf2025}. 
%The implementation of the layout was adopted from Kerbl et al. \cite{Kerbl2025}.
%The display concept features a horizontal split screen, adopted from the work of Kerbl et al. [source]. 
The upper half of the screen displays video streams from three front cameras. 
In contrast, the lower half shows a Lanelet2 map \cite{poggenhans2018lanelet2} with color-coded bounding boxes from the \ac{av}'s %(pedestrians: red, passenger vehicles: teal, larger vehicles: dark blue) 
\textit{Automation} vehicle interface (\cref{fig:architecture}), indicating object classes (e.g., teal for passenger vehicles). 
Furthermore, the map shows a LiDAR point cloud as ground truth information.
%\autoref{fig:trajectory_guidance} shows a disengagement due to the misclassification of road signs as pedestrians and larger vehicles. 
The bottom bar is displayed analogous to \textit{Direct Control} enhanced with additional \textit{Trajectory Guidance} relevant state information. %(yellow window).
%The bottom bar used in Direct Control mode, showing essential elements, including vehicle speed and latency information, is also used for Trajectory Guidance, with additional concept relevant state information.

    \section{Vehicle Platforms}
\label{sec:vehicles}

%The modular software architecture enables integration with different vehicles by adjusting vehicle parameters in the configuration files [ref to section?]. 
%The teleoperation concepts and software stack were tested on three vehicle platforms at the Technical University of Munich, as depicted in \cref{fig:visual_abstract}. 
The teleoperation software stack is demonstrated on three vehicle platforms, as depicted in \cref{fig:visual_abstract}. 

% The platforms address varying use cases.

To test and validate the software stack in a real-world environment, the research vehicle EDGAR is used.
EDGAR is equipped with several sensors, including cameras, LiDARs, and radar \cite{karle2024}. The automated driving software stack is based on Autoware. 
% real-world
%To ensure road safety, a safety driver is seated in the driver's seat and can intervene and take control in critical situations at any time.

% is used to develop and test the software on a real-world platform without the risks associated with operating a passenger vehicle, allowing teleoperation by non-experts. It is equipped with a LiDAR and high field-of-view cameras. The software runs on an NVIDIA Jetson.

Further, a virtual digital twin of EDGAR, referred to as vEDGAR, is utilized to prepare real-world testing and conduct user studies. vEDGAR integrates the Autoware software stack with CARLA as vehicle and environment simulation, enabling a full-stack simulation of the research vehicle.     
% Preparation for real world and studies

%For simulation, the existing simulation framework of the Institute of Automotive Technology, namely vEDGAR, is used for the EDGAR research vehicle in CARLA. 
%This approach ensures the vehicle interface remains consistent for both the actual vehicle and the simulation, enabling a closed-loop, full-stack simulation \cite{Gehrke2025}. 

The RoboRacer \cite{roboracer2025} enables real-world testing without the risks of using a full-sized vehicle, allowing teleoperation by non-experts. It is equipped with a LiDAR and high field-of-view cameras.
    \section{Experiments}
\label{sec:experiments}
%To demonstrate the overall functionality of the proposed teleoperation software, \textit{Direct Control} and \textit{Trajectory Guidance} are demonstrated using the EDGAR research vehicle in a common teleoperation scenario. Since latency is a key aspect for teleoperation and to provide a baseline for further research, a concise latency analysis of the most important data streams is carried out.
To demonstrate the overall functionality of the proposed teleoperation software, \textit{Direct Control} and \textit{Trajectory Guidance} are demonstrated on the platforms mentioned in \cref{sec:vehicles}. Since latency is a key aspect for teleoperation and to provide a baseline for further research, a concise latency analysis of the most important data streams is carried out.

\subsection{Latency Analysis}
%\todo[color=red!50]{@Nils Rework relative and gtg latency figures}
% \begin{itemize}
%     \item Describe which parts of the software introduce latency and differentiate between vehicle2operator (video streaming) and operator2vehicle (control command transmission) latency: figure?
% \end{itemize}

\subsubsection{Video Transmission}
The system is benchmarked on the EDGAR vehicle platform \cite{karle2024}, a modern high-end consumer PC and display setup. The PC is equipped with an Intel i9-14900KF CPU as well as an Nvidia RTX 4090 GPU and DDR5 RAM 4400\,\si{\mega\hertz}. The display runs at 240\,\si{\hertz} with an advertised Grey-to-grey reaction time of 0.03\,\si{\milli\second} and an overall input lag of about 3\,\si{\milli\second}. On the mentioned hardware, \textit{Visual} runs at a fixed frame rate of 240\,\si{\hertz}. 

To measure the glass-to-glass latency from the vehicle cameras to the \ac{ro}'s monitor, an Arduino setup with an LED and a photodiode that measures the delay between the last activation of the LED and the response of the photodiode attached to the display is used \cite{bachhuber_2016}. This setup follows previous work by Schimpe et al. \cite{schimpe2022} and Georg et al. \cite{georg_latency2020} using different hardware.
% \todo[]{Niklas: Does this need to be this detailed?}

\begin{figure}[!ht]
    \includegraphics[width=\linewidth]{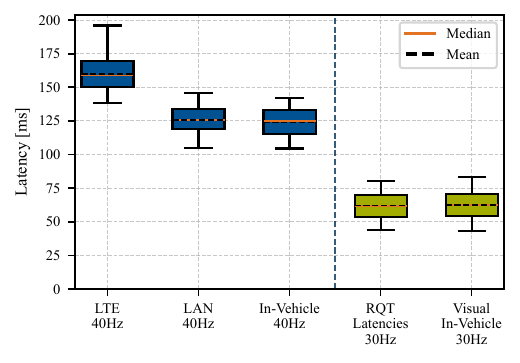}
    % \includesvg[width=\linewidth]{figures/g2g_figure_40Hz.svg}
    \caption{Glass-to-glass latency of the video streaming pipeline}
    \label{fig:g2g}
\end{figure}

To evaluate the latency, mid-range GigE cameras (Basler acA1920-50gc) of the tested platform are used \cite{karle2024}. The cameras were set to the target frame rates reported in \cref{fig:g2g}. For this, full tests including LAN (1\,\si{\giga\hertz}), LTE mobile network, as well as a in-vehicle connection and a directional connection to the cameras' ROS~2 driver are used. A measurement duration of 100\,\si{\second} was used. Each of these setups goes through the software's processing pipeline, including the encoding and decoding as well as network transmission via the RTSP stream, except for the in-vehicle test. In this case, \textit{Visual} is launched exclusively on the vehicle's compute platform. Further, the camera topic is directly subscribed to the camera driver's output.

The measured latencies range between 150\,\si{\milli\second} and 200\,\si{\milli\second} for the LTE transmission, with a median of 160\,ms at a camera framerate of 40\,\si{\hertz}. The observed in-vehicle latencies compared to the LAN transmission don't show a meaningful difference, both being around 125\,\si{\milli\second}. It can therefore be concluded that the induced network latency is around 35\,\si{\milli\second}. 

Directly subscribing to the image topic shows that the camera and rendering have a base latency of 60\,\si{\milli\second}. Thus, the rest of the video processing pipeline increases this latency by about 65\,\si{\milli\second} on average.

\subsubsection{Render Performance}
The render performance is carried out by comparing the rendering time of the software with other ROS~2 visualization software like rqt\_image\_view, Rviz2 as well as Foxglove-Studio. 
All applications are executed simultaneously on a single monitor with a refresh rate of 240\,\si{\hertz}, with the screen captured at the same refresh rate.
Each software subscribes to the same image topic, that alternates periodically between two colors. For each color switch, the number of frames since the updated color was rendered by the first application is recorded and converted into a relative rendering latency according to the monitor's refresh rate. The results of these rendering performance tests are summarized in \cref{tab:relative_latency}. On average, \textit{Visual} offers the shortest video rendering delay among the evaluated alternatives.

\begin{table}[h]
    \centering
    \caption{Average video rendering latency of different ROS~2 image visualization software relative to \textit{Visual}}
    \begin{tabular}{p{2.3cm} | p{2.3cm} | p{2.3cm}}
        \toprule
        rqt\_image\_view & Rviz2 & Foxglove-Studio \\
        \midrule
        + 16.6 \si{\milli\second} & + 29.2 \si{\milli\second} & + 37.5 \si{\milli\second} \\
        \bottomrule
    \end{tabular}
    \label{tab:relative_latency}
\end{table}

 %On average, rqt\_image\_view requires 16.6\,\si{\milli\second}, Rviz2 29.2\,\si{\milli\second}, and Foxglove-Studio  37.5\,\si{\milli\second} longer to render the same frame of the video.

% \begin{itemize}
%     \item Describe Arduino g2g test setup
%     \item Describe Network configurations: Same machine, in-vehicle network, LAN, mobile network 4G
%     \item Describe g2g results
%     \item Describe test setup for relative latency render test
%     \item Describe results for relative render latency test
% \end{itemize}

\subsubsection{Control Command Transmission}

In addition to video transmission and rendering, the transmission of control commands from \ac{ro} to the vehicle also introduces latency. So, the transmission latency towards the research platform EDGAR is measured as well as the overhead caused by the software. 

% \todo[color=red!30]{Xiyan: the recap can be deleted if necessary}

% As a recap from \ref{sec:network} and \ref{sec:direct_control}, the control command is defined as a custom ROS2 message and serialized into UDP or TCP packets before being transmitted through mobile network. The UDP/TCP packets are then deserialized into ROS2 messages in the vehicle and executed by the actuator. 

For the experiment, the same PC from the video transmission experiment is chosen. 
A NTP server is set up between the operator and the vehicle to synchronize the system clocks between the two computers. 
For the experiment, the \ac{ro} software is set to send a control command to the vehicle with a frequency of 1\,Hz. 
The tests are conducted through LTE mobile network and LAN. 

% add figure here
\begin{figure}[!ht]
    \includegraphics[width=\linewidth]{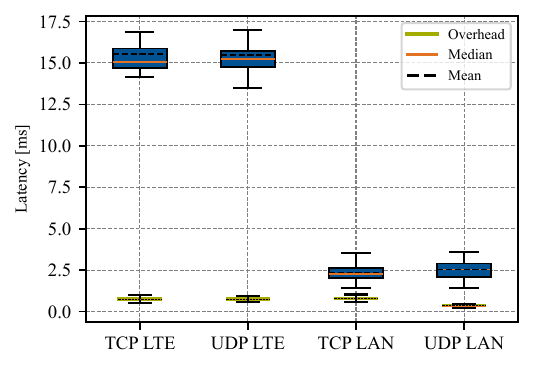}
    % \includesvg[width=\linewidth]{figures/images/ControlCmdLatency.svg}
    \caption{Transmission latency of control commands from the \ac{ro} to EDGAR through LTE and LAN using TCP/UDP}
    \label{fig:latency_control_cmd}
\end{figure}

The overhead, which includes the serialization and deserialization of the ROS~2 messages and the local transmission time of the \ac{dds} between ROS~2 topics, is less than 1 ms and takes up around 5\% of the overall latency for control command transmission over LTE for both UDP and TCP, as shown in Fig.~\ref{fig:latency_control_cmd}. The average transmission latency over LTE is $15.55 \pm 2.37$ ms for TCP and $15.49 \pm 1.81$ ms for UDP. 

\subsection{Qualitative System Performance}
\begin{figure}[!ht]
    \includegraphics[width=\linewidth]{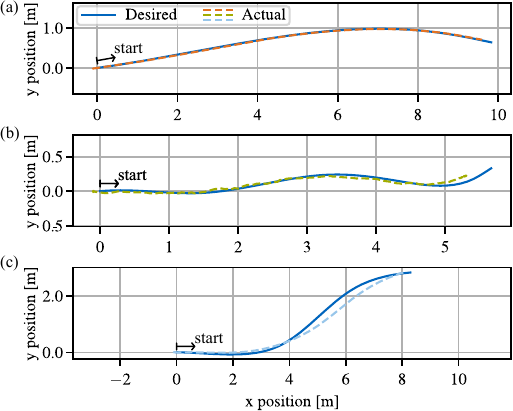}    
    \caption{Desired and actual path during \textit{Trajectory Guidance} for platforms EDGAR (a), RoboRacer (b) and vEDGAR (c)}
    \label{fig:path}
\end{figure}

Common scenarios for teleoperation of \acp{av} include navigating lane blockages or alterations due to construction sites \cite{Brecht2024EvaluationOfConcepts}.
The software's functionality is demonstrated by performing a scenario on each of the three platforms mentioned in \cref{sec:vehicles}. For this, the \ac{ro} should bypass a construction site using both \textit{Direct Control} and \textit{Trajectory Guidance}. The requested steering angles are executed instantaneously by the platform, an offset in the simulation internal steering wheel calculation can however be observed. A precise path execution for \textit{Trajectory Guidance} is visible in the provided \cref{fig:path}.

\begin{figure}[!ht]
    \includegraphics[width=\linewidth]{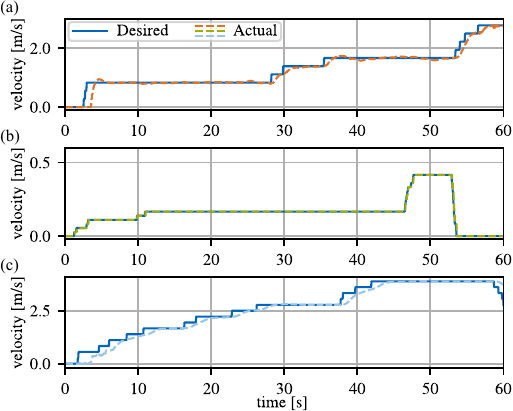}
    \caption{Desired and actual velocity during \textit{Direct Control} for platforms EDGAR (a), RoboRacer (b) and vEDGAR (c)}
    \label{fig:velocity}
\end{figure}

\begin{figure}[!ht]
    \includegraphics[width=\linewidth]{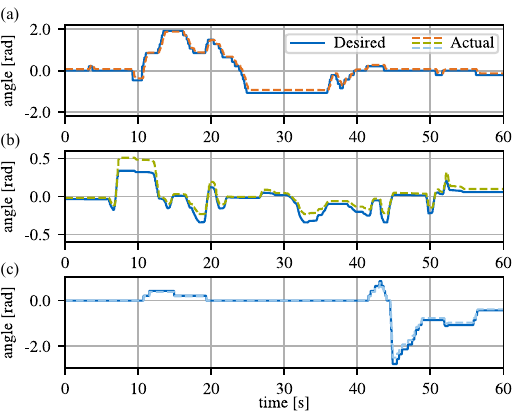}
    \caption{Desired and actual steering wheel angle during \textit{Direct Control} for platforms EDGAR (a), RoboRacer (b) and vEDGAR (c)}
    \label{fig:steering}
\end{figure}

\cref{fig:velocity} depicts the velocity tracking performance in \textit{Direct Control} mode. The results show a stable stationary behavior but slow response for platforms EDGAR and vEDGAR.  
The tracking of the second operator input, steering angle for \textit{Direct Control} and path for \textit{Trajectory Guidance}, is shown in \cref{fig:steering} and \cref{fig:path}, respectively. 

%A video showcasing the usage of the software in the munich inner-city is available online\footnote[1]{https://youtu.be/...}. \textcolor{red}{Provide Video}
    \section{Conclusion}
\label{sec:conclusion}
In this work, a teleoperation software is introduced, designed by requirements derived upon current research activities in the field. A modular and interchangeable software architecture allows a simple integration of components and extension of the software stack. Thus, new approaches for teleoperation concepts, especially \ac{ra} concepts, can now be easily implemented and compared to existing concepts. Additionally, it enables further development and research on user interfaces and key components of the software, e.g., the network. With its generic interfaces, different platforms and automated driving software stacks like Autoware \cite{git_autoware} can be used with the proposed software stack. The overall functionality of the software is demonstrated in a simulation environment, on a small-scaled vehicle as well as on a research vehicle by using the two teleoperation concepts \textit{Direct Control} and \textit{Trajectory Guidance}. In addition, a comprehensive latency analysis provides a baseline for further development.

Future work will focus on community-driven development of the software to extend its capabilities and encourage the usage of the software for research. Building on the presented software, teleoperation, safety, and user interface concepts, as well as %core 
functionalities like video streaming and networking, will be integrated and evaluated through user studies to further advance research in the field of teleoperation.

% LONG VERSION
%Future work will examine different safety concepts, as well as the integration into a control center framework. The introduced modularity and exchangeability allows to extend the software by further teleoperation concepts. Additionally, the available code base enables advanced research and studies in the field of teleoperation. It supports the assessment of interaction with both \ac{ro} and automated driving systems. Furthermore, data transmission over mobile networks can be evaluated and enhanced within the teleoperation stack under real-world conditions.

% SHORT VERSION
%Future Work will examine advanced research and studies in the field of teleoperation. The introduced modularity and exchangeability allows to extend the software by further teleoperation, safety and HMI concepts. 

    %% After
     \section*{Acknowledgment}
Tobias Kerbl, David Brecht, Nils Gehrke, Nijinshan Karunainayagam, Niklas Krauss, Florian Pfab,
Richard Taupitz, Ines Trautmannsheimer, Xiyan Su and Maria-Magdalena Wolf, as the first authors, collectively contributed to the presented work. Frank Diermeyer made essential contributions to the conception of the research projects and revised the paper critically. 
He gave final approval for the version to be published and agrees to all aspects of the work. As a guarantor, he accepts responsibility for the overall integrity of the paper. 
The authors want to thank Gemb Kaljavesi for providing the tool used in the rendering performance experiments.
This work was supported by the Federal Ministry of Education and Research of Germany within the projects AUTOtech.agil (FKZ 01IS22088) and ConnRAD (FKZ 16KISR034), the project Wies’n Shuttle (FKZ 03ZU1105AA) in the MCube cluster, the Federal Ministry of Economic Affairs and Climate Actions of Germany within the projects ATLAS-L4 (FKZ 19A21048l) and Safestream (FKZ 01ME21007B), and through basic research funds from the Institute for Automotive Technology.
    \begin{acronym}
    \acro{ros}[ROS]{Robot Operating System}
    \acro{tc}[TC]{Teleoperation Concept}
    \acroplural{tc}[TCs]{Teleoperation Concepts}
    \acro{av}[AV]{Automated Vehicle}
    \acroplural{av}[AVs]{Automated Vehicles}
    \acro{edgar}[EDGAR]{Excellent Driving GARching}
    \acro{gui}[GUI]{Graphical User Interface}
    \acro{rd}[RD]{Remote Driving}
    \acro{ra}[RA]{Remote Assistance}
    \acro{ro}[RO]{Remote Operator}
    \acroplural{ro}[ROs]{Remote Operators}
    \acro{odd}[ODD]{Operational Design Domain}
    \acro{srtp}[SRTP]{Successive Reference Pose Tracking}
    \acro{kpi}[KPI]{Key-Performance-Indicator}
    %\acro{DDT}[DDT]{Dynamic Driving Task}           % streichen
    \acro{ttc}[TTC]{Time to Collision}
    \acro{gui}[GUI]{Graphical User Interface} 
    \acro{hmi}[HMI]{Human Machine Interface} 
    \acro{sae}[SAE]{Society of Automotive Engineers}
    \acro{dds}[DDS]{Data Distribution Service}
    \acro{ads}[ADS]{Automated Driving System}
\end{acronym}

    \bibliographystyle{IEEEtran}
    \bibliography{IEEEabrv,bib}

\end{document}